\documentclass[11pt]{article}

\usepackage[preprint]{acl}

\usepackage{times}
\usepackage{latexsym}

\usepackage[T1]{fontenc}
\usepackage[utf8]{inputenc}

\usepackage{microtype}
\usepackage{inconsolata}

\usepackage{graphicx}
\usepackage{amsmath,amssymb}
\usepackage{booktabs}
\usepackage{multirow}
\usepackage{url}

\title{A Large Language Model Based Method for Complex Logical Reasoning over Knowledge Graphs}

\author{Ziyan Zhang\textsuperscript{1}, Chao Wang\textsuperscript{2}, Zhuo Chen\textsuperscript{2}, Lei Chen\textsuperscript{2}, Chiyi Li\textsuperscript{2}, Kai Song\textsuperscript{1} \\
$^1$School of Information Science and Engineering, Chongqing Jiaotong University\\ $^2$State Grid Chongqing Electric Power Company\\
}

\begin{document}
\maketitle

\begin{abstract}
Reasoning over knowledge graphs (KGs) with first-order logic (FOL) queries is challenging due to the inherent incompleteness of real-world KGs and the compositional complexity of logical query structures.
Most existing methods rely on embedding entities and relations into continuous geometric spaces and answer queries via differentiable set operations. While effective for simple query patterns, these approaches often struggle to generalize to complex queries involving multiple operators, deeper reasoning chains, or heterogeneous KG schemas.
We propose ROG (Reasoning Over knowledge Graphs with large language models), an ensemble-style framework that combines query-aware KG neighborhood retrieval with large language model (LLM)-based chain-of-thought reasoning. ROG decomposes complex FOL queries into sequences of simpler sub-queries, retrieves compact, query-relevant subgraphs as contextual evidence, and performs step-by-step logical inference using an LLM, avoiding the need for task-specific embedding optimization.
Experiments on standard KG reasoning benchmarks demonstrate that ROG consistently outperforms strong embedding-based baselines in terms of mean reciprocal rank (MRR), with particularly notable gains on high-complexity query types. These results suggest that integrating structured KG retrieval with LLM-driven logical reasoning offers a robust and effective alternative for complex KG reasoning tasks.
\end{abstract}

\section{Introduction}
Knowledge graphs (KGs) store and encode structured knowledge in the form of networks composed of relational triples.
Entity nodes in such networks are connected via edges representing relations between entities, as exemplified by large-scale knowledge bases such as Freebase \citep{bollacker2008freebase} and WordNet \citep{miller1995wordnet}.
However, real-world knowledge graphs such as Freebase are typically characterized by large scale, substantial noise, and inherent incompleteness.
As a result, performing reasoning over KGs has become a challenging problem in artificial intelligence research.
In particular, answering first-order logic (FOL) queries \citep{enderton2001mathematical} over KGs is widely regarded as a representative and difficult task in this domain \citep{liang2024kg}.
FOL queries over knowledge graphs are usually defined using existential quantification ($\exists$) together with a small set of logical operators, including intersection ($\wedge$), projection ($P$), union ($\vee$), and negation ($\neg$).
By composing these operators, complex logical constraints over entities and relations can be expressed.

\begin{figure*}[t]
  \centering
  \includegraphics[width=\linewidth]{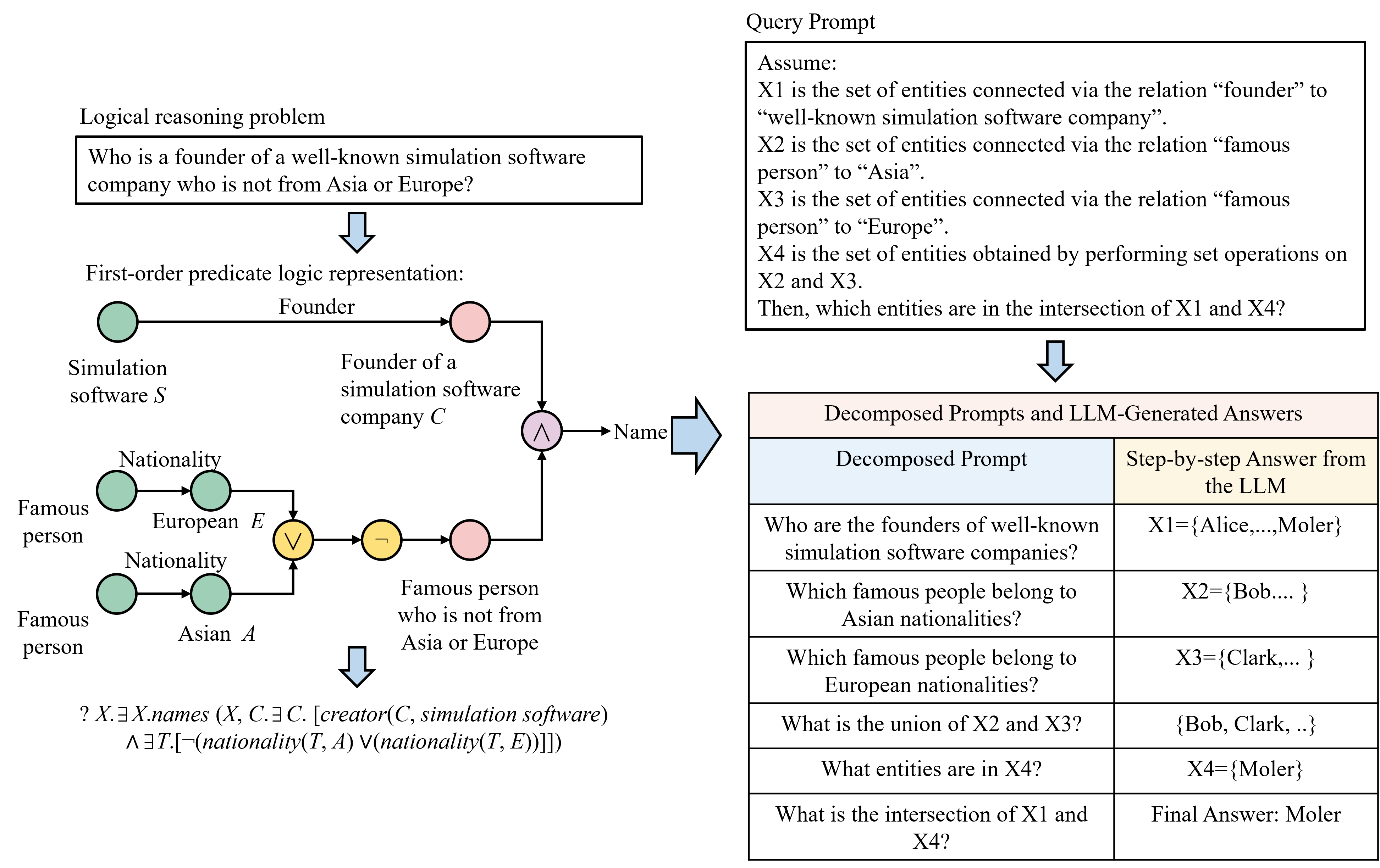}
  \caption{Workflow of Logical Reasoning Decomposition and LLM-Based Answer Generation.}
  \label{fig:fig1}
\end{figure*}

Existing research on this topic has mainly focused on capturing the semantic positions of entities and the logical coverage of queries through various forms of \emph{geometric embedding} methods \citep{hamilton2018embedding}.
Although such approaches are effective in many cases, they still suffer from several notable limitations in practice.
First, the reasoning capability of traditional methods is often limited when dealing with complex logical queries \citep{choudhary2023complex}.
Due to the constraints imposed by formalized FOL query representations, chaining multiple reasoning steps or modeling rich multi-relational dependencies among entities can easily lead to loss of information and degraded performance  \citep{choudhary2022probabilistic}.
Second, these methods typically exhibit limited generalization ability \citep{teru2020inductive}.
Models optimized for a specific KG or query distribution often fail to generalize effectively to other heterogeneous KGs, which restricts their applicability across different domains and schemas.
To address the above limitations, this work leverages the reasoning capabilities of large language models (LLMs) and proposes an \emph{ensemble-style} reasoning paradigm for complex logical reasoning over knowledge graphs.
Specifically, we introduce a natural language model–guided knowledge graph reasoning framework based on LLM inference.

Figure~\ref{fig:fig1} illustrates an example of query-chain decomposition and LLM-based logical inference in our framework, referred to as \textbf{ROG} (LLM reasoning over knowledge graphs).
Since LLMs are particularly effective at answering simple queries, our approach decomposes complex logical queries involving multiple operators into a sequence of basic queries, each containing only a single operator.
The final answer is then obtained by applying an LLM-driven sequential answering mechanism over these decomposed queries.

Concretely, ROG retrieves query-relevant subgraph context from the KG using logical query constraints, and performs chain-style inference over the retrieved context using carefully designed LLM prompts \citep{huang2024trustworthy}.
The framework first abstracts logical information from the input query and the KG, allowing ROG to focus on logical structure rather than surface semantics.
By explicitly separating logical representation from natural language generation, ROG mitigates hallucination effects and improves reasoning robustness.
From the abstracted KG, entities and relations involved in logical queries are extracted to construct query-relevant subgraphs, which are then provided as contextual input to the LLM.
Subsequently, complex logical queries containing multiple operators are deterministically decomposed into basic queries following logical order.
Each basic query is processed independently, and intermediate answers are stored in a cache for efficient access during subsequent reasoning steps.

This strategy effectively integrates logical reasoning over knowledge graphs with the inference capabilities of large language models, resulting in a novel and flexible reasoning framework.
Compared with traditional methods that are constrained by explicit FOL representations, ROG can be viewed as a KG-enhanced reasoning approach that fully exploits LLM inference capability.
Overall, the main contributions of this work can be summarized as follows:
\begin{itemize}
  \item We propose a novel LLM-based framework, ROG, for complex logical reasoning over knowledge graphs, which leverages LLM inference capabilities to efficiently answer FOL queries over KGs.
  \item ROG retrieves query-relevant subgraph context by identifying entities and relations involved in logical queries, and performs chain-style reasoning over these subgraphs using decomposed query prompts.
  \item Experiments on standard KG benchmarks demonstrate that, across 14 FOL query types involving projection ($P$), intersection ($\wedge$), union ($\vee$), and negation ($\neg$), ROG achieves performance improvements of approximately 35\%--55\% over prior methods.
\end{itemize}

\section{Related Work}

Our work is closely related to two research directions: prompt-based reasoning with large language models, and logical reasoning over knowledge graphs.

\paragraph{Prompt-based reasoning with large language models.}
Recent studies have shown that large language models (LLMs) can perform a wide range of natural language processing (NLP) tasks by simply providing contextual prompts, without requiring task-specific fine-tuning \citep{wu2024power}.
Beyond single-step prediction, LLMs have also been successfully applied to multi-step reasoning tasks by explicitly providing intermediate reasoning steps, commonly referred to as chain-of-thought (CoT) prompting \citep{wei2022chain}.
Some approaches \citep{creswell2022selection, gao2023pal, yang2024harnessing, yang2024can, xiong2025deliberate} further combine multiple LLMs, or LLMs with symbolic components, and predefine decomposition structures that guide the execution of multi-step reasoning processes.

For example, sequential and decompositional prompting strategies \citep{zhou2022least, khot2022decomposed, wang2022self, yao2023tree} break complex prompts into a series of simpler prompts and require the model to answer them in a predefined order, thereby enabling complex reasoning.
Although these methods are conceptually related to our approach, they typically do not leverage intermediate answers to guide subsequent reasoning steps.
In contrast, the proposed ROG framework explicitly incorporates logical structure into a chain-style decomposition mechanism, enhances reasoning by retrieving relevant KG neighborhoods, and adopts a multi-stage answering process in which intermediate LLM outputs are cached and reused in subsequent queries.
This design allows ROG to fuse previous LLM answers into later reasoning stages, enabling more coherent and logically consistent inference.

\paragraph{Logical reasoning over knowledge graphs.}
Early work on logical reasoning over knowledge graphs primarily focused on capturing semantic information of entities and modeling relational operators involved in projection queries \citep{nickel2011rescal, bordes2013translating}.
Subsequent studies revealed the need to encode spatial and hierarchical information inherent in KGs using geometric representations \citep{nickel2017poincare}.
To address this challenge, Query2Box \citep{ren2020query2box} and BetaE \citep{ren2020beta} represent entities and relations as box embeddings and probability distributions, respectively, enabling the modeling of uncertainty and set-based logical operations.

Other approaches, such as CQD \citep{minervini2021complex}, focus on improving complex logical reasoning performance by combining answers from simpler intermediate queries.
In addition, QA-GNN \citep{yasunaga2021qagnn} enhances question answering performance by integrating KG neighborhood information with neural reasoning models.
Recent work \citep{xiongtilp, xiong2024teilp, xiong2024large} has further extended knowledge graphs to temporal knowledge graphs in order to incorporate temporal information.
Overall, prior work in this area has largely concentrated on improving KG representations and embedding-based reasoning architectures or logical rules to support logical inference.

In contrast to these approaches, we propose a system-level framework that directly leverages the reasoning capabilities of large language models and tailors them to logical reasoning tasks over knowledge graphs.
Rather than designing task-specific embedding spaces for logical operators, ROG integrates KG neighborhood retrieval with LLM-driven chain-of-thought reasoning, offering a flexible and general alternative for complex KG reasoning problems.

\begin{figure*}[t]
  \centering
  \includegraphics[width=\linewidth]{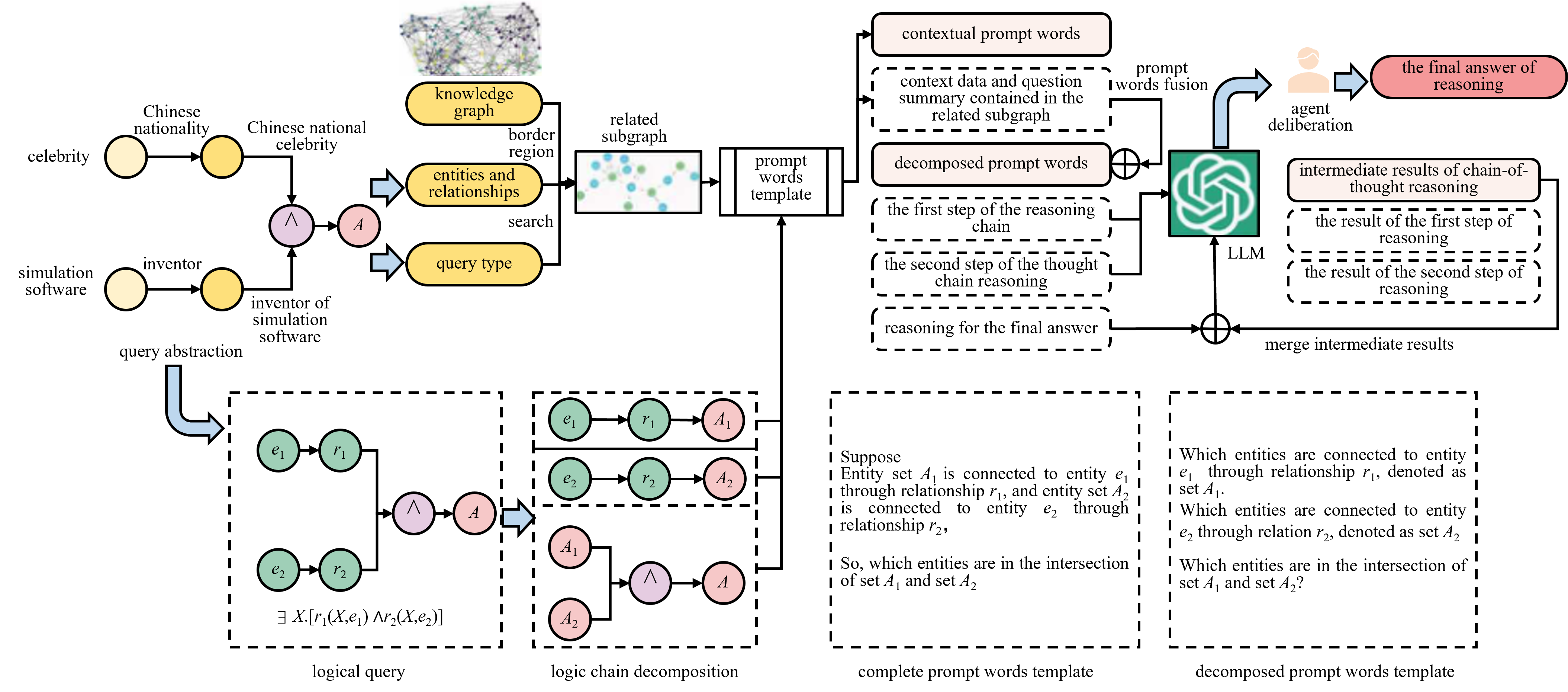}
  \caption{Data flow of the ROG model.}
  \label{fig:fig2}
\end{figure*}

\section{Methodology}

\subsection{Formalization}
We consider logical reasoning over a knowledge graph (KG)
$G \subseteq E \times R \times E$,
where $E$ denotes the entity set and $R$ denotes the relation set.
A KG can be represented as a collection of relational triples
$\langle e_1, r, e_2 \rangle \in G$,
where each relation $r \in R$ is a Boolean function
$r : E \times E \rightarrow \{\texttt{True}, \texttt{False}\}$,
indicating whether two entities are connected by relation $r$.

We focus on four basic first-order logic (FOL) operators commonly used in KG reasoning:
projection ($P$), intersection ($\wedge$), union ($\vee$), and negation ($\neg$).
Based on these operators, logical queries over the KG can be formally defined as follows:
\begin{align}
q_P &= P_v : \{v_1,\ldots,v_k\} \subseteq E \;\exists a_1, \label{eq:qP} \\
q_\wedge &= \wedge_v : \{v_1,\ldots,v_k\} \subseteq E \;\exists a_1 \wedge \cdots \wedge a_i, \label{eq:qAnd} \\
q_\vee &= \vee_v : \{v_1,\ldots,v_k\} \subseteq E \;\exists a_1 \vee \cdots \vee a_i, \label{eq:qOr} \\
q_{\neg} &= \neg_v : \{v_1,\ldots,v_k\} \subseteq E \;\exists \neg a_1. \label{eq:qNot}
\end{align}

Here, $a_i = r_i(e_i, v_i)$ denotes an adjacency indicator:
$a_i = 1$ if entity $e_i$ is connected to $v_i$ through relation $r_i$, and $a_i = 0$ otherwise.
Among these query types, $P$, $\wedge$, $\vee$, and $\neg$ are referred to as projection, intersection, union, and negation queries, respectively.
Each query takes a set of entities and relations $\{e_i, r_i\}$ as input, which specify the logical constraints of the query.
The goal of logical reasoning is to apply these operators to the KG and output the corresponding query results.

\subsection{Neighborhood Retrieval and Query-Chain Decomposition}
ROG is built on top of a large language model (LLM).
Due to context-length limitations, an LLM cannot directly process the entire knowledge graph.
However, logical reasoning primarily depends on relations between entities rather than the full graph structure.
Therefore, we design a prompt-based reasoning framework that enables the LLM to perform logical inference over a restricted, query-relevant subgraph.

\paragraph{Query abstraction.}
We replace concrete entity and relation names in the KG and queries with unique identifiers (IDs).
By doing so, ROG avoids reliance on the LLM's external world knowledge and reduces hallucination, while improving generalization across different KGs.

\paragraph{Neighborhood retrieval.}
To enable efficient reasoning without accessing the full KG, we retrieve a query-relevant $k$-hop neighborhood.
Let $E_q$ and $R_q$ denote the sets of entities and relations involved in query $q$.
We define the neighborhood $N_k(q)$ recursively as:
\begin{align}
N_1(q) &= \{(e,r) : (e \in E_q) \wedge (r \in R_q)\}, \label{eq:N1} \\
E_q^{k} &= \{e : (e,r) \in N_{k-1}(q)\}, \label{eq:Ek} \\
R_q^{k} &= \{r : (e,r) \in N_{k-1}(q)\}, \label{eq:Rk} \\
N_k(q) &= \{(e,r) : (e \in E_q^{k}) \wedge (r \in R_q^{k})\}. \label{eq:Nk}
\end{align}

The computational complexity of a query is proportional to the number of entities and relations involved, i.e.,
$O(q) = |E_q| + |R_q|$.
To reduce complexity, ROG decomposes a complex multi-operator query into multiple single-operator sub-queries.

For example, a $k$-hop projection query ($kp$) can be decomposed into $k$ consecutive single-hop projection queries.
Specifically, a $3p$ query involving one entity and three relations,
$e_1 \xrightarrow{r_1} \xrightarrow{r_2} \xrightarrow{r_3} A$,
can be decomposed as:
$e_1 \xrightarrow{r_1} A_1$,
$A_1 \xrightarrow{r_2} A_2$,
$A_2 \xrightarrow{r_3} A$,
where $k=3$ is determined by the query type.

Similarly, an intersection query can be decomposed into multiple projection queries followed by a final intersection.
For instance, a $3i$ query
$(e_1 \xrightarrow{r_1}) \wedge (e_2 \xrightarrow{r_2}) \wedge (e_3 \xrightarrow{r_3}) = A$
is decomposed into
$e_1 \xrightarrow{r_1} A_1$,
$e_2 \xrightarrow{r_2} A_2$,
$e_3 \xrightarrow{r_3} A_3$,
followed by
$A = A_1 \wedge A_2 \wedge A_3$.
Other query types can be decomposed in a similar manner.

\begin{table*}[t]
\centering
\small
\begin{tabular}{llccccccccc}
\toprule
Dataset & Model & 1p & 2p & 3p & 2i & 3i & ip & pi & 2u & up \\
\midrule
\multirow{4}{*}{FB15k}
& GQE & 57.2 & 55.9 & 29.9 & 52.4 & 47.7 & 41.8 & 44.5 & 29.8 & 31.2 \\
& Q2B & 58.1 & 48.7 & 43.3 & 65.1 & 56.9 & 47.1 & 50.2 & 35.2 & 29.9 \\
& CQD & 72.4 & 56.9 & 41.5 & 64.6 & 61.3 & 54.4 & 58.8 & 55.7 & 36.9 \\
& ROG & \textbf{81.4} & \textbf{67.7} & \textbf{49.2} & \textbf{75.6} & \textbf{72.3} & \textbf{62.0} & \textbf{65.1} & \textbf{69.4} & \textbf{45.6} \\
\midrule
\multirow{4}{*}{NELL995}
& GQE & 52.5 & 29.5 & 15.3 & 35.6 & 38.8 & 28.5 & 25.4 & 26.9 & 18.6 \\
& Q2B & 53.1 & 31.0 & 22.4 & 51.4 & 46.9 & 31.1 & 29.1 & 37.6 & 21.5 \\
& CQD & 66.4 & 32.7 & 26.8 & 55.8 & 51.2 & 37.7 & 33.6 & 45.4 & 39.2 \\
& ROG & \textbf{83.3} & \textbf{59.2} & \textbf{41.5} & \textbf{61.1} & \textbf{58.7} & \textbf{42.9} & \textbf{50.1} & \textbf{62.2} & \textbf{57.2} \\
\bottomrule
\end{tabular}
\caption{MRR comparison between ROG and traditional methods on nine typical query types.}
\label{tab:typical}
\end{table*}

\subsection{Chain-of-Thought Prompting}
After decomposing a complex query into a chain of simple sub-queries, ROG leverages the reasoning capability of the LLM to sequentially generate query results.
To this end, we design prompt templates that convert structured data into natural language prompts and feed them to the LLM step by step.

Specifically, the retrieved subgraph is transformed into contextual prompts, and each decomposed sub-query is converted into a corresponding reasoning prompt.
Intermediate results produced by the LLM are stored in a cache and reused in subsequent reasoning steps.
During prompt construction, placeholder tokens are employed to insert intermediate answers into later prompts, ensuring the completeness and consistency of the reasoning chain.

Furthermore, ROG introduces an agent-based mechanism for reasoning in domain-specific scenarios.
An intelligent agent is defined as a combination of a knowledge base and an LLM.
Multiple agents collaborate through a ``consensus'' mechanism to ensure that the generated answers are interpretable and logically consistent.
This design further enhances the robustness and explainability of ROG's reasoning process.

\section{Experiments}

All experiments are conducted on a machine equipped with an NVIDIA GeForce RTX 4090 GPU.
We use the ChatGLM series \citep{glm2024chatglm} and construct knowledge graphs based on Neo4j.
Two sets of comparative experiments are performed.
The results demonstrate that the proposed ROG framework consistently outperforms traditional methods on logical reasoning tasks.
Moreover, the reasoning paradigm based on chain-of-thought query decomposition and logical answer ordering is shown to be practically effective.

\subsection{Datasets and Baselines}

We evaluate the proposed ROG framework on two widely used public benchmarks: FB15k \citep{bordes2013translating} and NELL995 \citep{xiong2017deeppath}.
The following representative baselines are considered for comparison:

\begin{itemize}
    \item \textbf{GQE} encodes each logical query into a single vector and represents entities and relations in a low-dimensional space. Projection and intersection operations are modeled using transformation operators and deep set operators, respectively.
    \item \textbf{Query2Box (Q2B)} adopts box embeddings, which generalize conventional vector embeddings and are capable of capturing richer semantic regions.
    \item \textbf{CQD} decomposes complex logical queries into simpler sub-queries and applies query-specific attention mechanisms to combine intermediate results.
\end{itemize}

\begin{table}[t]
\centering
\small
\begin{tabular}{llccccc}
\toprule
Dataset & Model & 2in & 3in & inp & pin & pni \\
\midrule
\multirow{2}{*}{FB15k}
& BetaE & 29.8 & 21.7 & 19.1 & 14.4 & 18.4 \\
& ROG   & \textbf{36.4} & \textbf{35.7} & \textbf{22.3} & \textbf{19.6} & \textbf{25.0} \\
\midrule
\multirow{2}{*}{NELL995}
& BetaE & 29.6 & 27.9 & 23.4 & 21.1 & 15.6 \\
& ROG   & \textbf{34.7} & \textbf{32.8} & \textbf{29.5} & \textbf{27.3} & \textbf{19.5} \\
\bottomrule
\end{tabular}
\caption{MRR comparison between ROG and BetaE on complex query types.}
\label{tab:complex}
\end{table}

\subsection{Effectiveness of ROG for Logical Reasoning}

We evaluate ROG by executing a series of standard logical queries and comparing it against the baseline methods.

\paragraph{Multi-hop Projection Queries.}
These queries project from a starting entity to target entities via multiple relational hops.
We consider one-hop, two-hop, and three-hop projection queries, denoted as \textit{1p}, \textit{2p}, and \textit{3p}, respectively.

\paragraph{Intersection and Union Queries.}
We evaluate queries involving intersection ($\wedge$) and union ($\vee$) operations.
Specifically, we test intersection queries over two and three entity sets, denoted as \textit{2i} and \textit{3i}, and union queries over two entity sets, denoted as \textit{2u}.

\paragraph{Composite Queries.}
To simulate realistic reasoning over knowledge graphs, we evaluate composite queries that combine multiple logical operators, including projection, intersection, and union.

\paragraph{Negation Queries.}
We further evaluate queries involving negation.
Specifically, we consider negation applied to \textit{2i}, \textit{3i}, \textit{ip} (intersection followed by projection), and \textit{pi} (projection followed by intersection), denoted as \textit{2in}, \textit{3in}, \textit{inp}, and \textit{pin}, respectively.
We also analyze \textit{pni}, where negation covers two entities in the intersection.

\paragraph{Evaluation Metric.}
We adopt Mean Reciprocal Rank (MRR) to evaluate reasoning quality.
MRR measures how close the retrieved result is to the optimal answer, with larger values indicating better performance.
It is defined as:
\begin{equation}
\mathrm{MRR} = \frac{1}{N} \sum_{i=1}^{N} \frac{1}{p_i},
\end{equation}
where $N$ denotes the number of queries and $p_i$ is the rank of the correct answer in the candidate list for the $i$-th query.

Table~\ref{tab:typical} reports the MRR scores of four reasoning methods evaluated on two datasets across nine types of logical queries.
Here, $p$, $i$, and $u$ denote projection, intersection, and union operations, respectively, while the numerical prefixes indicate the query depth or the number of involved entity sets.

Overall, ROG consistently achieves higher MRR scores across all query types.
Compared with traditional methods, ROG improves MRR by approximately 35\%--55\% on average.

For queries involving projection operations, the effectiveness of ROG generally improves as the reasoning depth increases, although its absolute performance may still be slightly lower than that of traditional methods on shallow queries.
For queries involving intersection and union operations, ROG exhibits stronger robustness and maintains stable performance, while traditional methods tend to degrade more noticeably.
Furthermore, when considering queries that combine projection, intersection, and union, we observe that as query depth increases, ROG adapts better to the growth in query breadth, demonstrating improved scalability with respect to complex query structures.

\subsection{Advantages of Chain-of-Thought Decomposition}
\label{sec:cot_advantage}

The goal of this experiment is to investigate the advantages of ROG’s chain-of-thought (CoT) decomposition strategy compared with traditional complex query reasoning methods.
Embedding-based approaches such as BetaE commonly address multi-hop logical reasoning in knowledge graphs by modeling path search and embedding transformations.
In BetaE, the embedding of an entity is treated as a Beta distribution, and the embeddings of intermediate reasoning steps are iteratively updated through distribution transformations.

In contrast, the results in Table~\ref{tab:complex} show that ROG significantly improves performance by approximately 25\%--35\% on complex queries.
These results clearly indicate that LLMs are capable of capturing the broad logical semantics required for complex reasoning.
By decomposing complex queries into sequences of simpler sub-queries, ROG effectively enhances reasoning performance.
This study therefore demonstrates the potential of chain-of-thought decomposition to overcome the limitations of traditional complex query reasoning methods and to improve the efficiency of logical reasoning tasks over knowledge graphs.

\section{Conclusion}

We presented ROG, a framework that integrates logical reasoning over knowledge graphs with the inference capabilities of large language models.
By decomposing complex first-order logic queries into sequences of simpler sub-queries and performing chain-of-thought reasoning over query-relevant subgraphs, ROG achieves consistent performance improvements over traditional methods on standard KG benchmarks.
The results suggest that combining structured KG retrieval with LLM-guided reasoning provides an effective and flexible solution for complex logical reasoning over large, noisy, and incomplete knowledge graphs.

\section*{Limitations}

ROG depends on the reasoning capability and prompt stability of the underlying large language model, and its performance may vary across different model choices.
Although query-relevant subgraph retrieval alleviates context-length constraints, retrieval errors or incomplete neighborhoods can still affect downstream reasoning quality.
Moreover, the current evaluation is limited to standard KG benchmarks with predefined query structures, and scaling to larger or more open-ended settings may require more efficient retrieval and context management.

\bibliography{custom}

\end{document}